\documentclass[letterpaper]{article} 
\usepackage{aaai25}  
\usepackage{times}  
\usepackage{helvet}  
\usepackage{courier}  
\usepackage[hyphens]{url}  
\usepackage{graphicx} 
\usepackage{natbib}  
\usepackage{caption} 
\usepackage{algorithm}
\usepackage{algorithmic}
\usepackage{url}
\usepackage{newfloat}
\usepackage{listings}
\usepackage{amsmath}
\usepackage{array}      
\usepackage{booktabs}   
\usepackage[table]{xcolor}   
\usepackage{colortbl}   
\usepackage{makecell}   
\usepackage{amssymb}    
\usepackage{wasysym}
\usepackage{multicol}
\usepackage{multirow}
\usepackage{diagbox}
\usepackage{gensymb}
\usepackage{tabularx}
\usepackage{subfigure}

\def\ie{\textit{i.e.}}
\def\eg{\textit{e.g.}}

\DeclareCaptionStyle{ruled}{labelfont=normalfont,labelsep=colon,strut=off} 
\floatstyle{ruled}
\newfloat{listing}{tb}{lst}{}
\floatname{listing}{Listing}
\title{Breaking Barriers in Physical-World Adversarial Examples: Improving Robustness and Transferability via Robust Feature}
\author{
    Yichen Wang\textsuperscript{\rm 1,2,4,5,$*$}, 
    Yuxuan Chou\textsuperscript{\rm $*$}, 
    Ziqi Zhou\textsuperscript{\rm 1,2,3,$\dagger$}, 
    Hangtao Zhang\textsuperscript{\rm $*$},
    Wei Wan\textsuperscript{\rm 1,2,4,5,$*$}, \\
    Shengshan Hu\textsuperscript{\rm 1,2,4,5,$*$},
    Minghui Li\textsuperscript{\rm $\ddagger$}
}
\affiliations{
    \textsuperscript{\rm 1}National Engineering Research Center for Big Data Technology and System\\
    \textsuperscript{\rm 2}Services Computing Technology and System Lab
    \textsuperscript{\rm 3}Cluster and Grid Computing Lab\\
    \textsuperscript{\rm 4}Hubei Engineering Research Center on Big Data Security \\
    \textsuperscript{\rm 5}Hubei Key Laboratory of Distributed System Security\\
     $*$ School of Cyber Science and Engineering,
Huazhong University of Science and Technology\\
 $\dagger$ School of Computer Science and Technology, 
Huazhong University of Science and Technology \\
$\ddagger$ School of School of Software Engineering, Huazhong University of Science and Technology \\

    \{wangyichen,yuxuanchou,zhouziqi, hangt\_zhang,wanwei\_0303, hushengshan, minghuili\}@hust.edu.cn

}

\usepackage{bibentry}

\begin{document}

\maketitle

\begin{abstract}
As deep neural networks (DNNs) are widely applied in the physical world, many researches are focusing on physical-world adversarial examples (PAEs), which introduce perturbations to inputs and cause the model's incorrect outputs. 
However, existing PAEs face two challenges: 
{unsatisfactory attack performance (\ie, poor transferability and insufficient robustness to environment conditions)}, and difficulty in balancing attack effectiveness with stealthiness, where better attack effectiveness often makes PAEs more perceptible.

In this paper, we explore a novel perturbation-based method to overcome the challenges. {For the first challenge, we introduce a strategy Deceptive RF injection based on robust features (RFs) that are predictive, robust to perturbations, and consistent across different models. Specifically, it improves the transferability and robustness of PAEs by covering RFs of other classes onto the predictive features in clean images.} 
For the second challenge, we introduce another strategy Adversarial Semantic Pattern Minimization, which removes most perturbations and retains only essential adversarial patterns in AEs. 
Based on the two strategies, we design our method Robust Feature Coverage Attack (RFCoA), comprising Robust Feature Disentanglement and Adversarial Feature Fusion. In the first stage, we extract target class RFs in feature space. In the second stage, we use attention-based feature fusion to overlay these RFs onto predictive features of clean images and remove unnecessary perturbations. Experiments show our method's superior transferability, robustness, and stealthiness compared to existing state-of-the-art methods.  Additionally, our method's effectiveness can extend to Large Vision-Language Models (LVLMs), indicating its potential applicability to more complex tasks.

\end{abstract}

%
\begin{links}
    \link{Code}{https://github.com/CGCL-codes/RFCoA}
\end{links}

\section{Introduction}
Deep neural networks (DNNs) have achieved significant milestones in various domains such as image recognition \cite{densenet}, natural language processing \cite{gpt4}, and speech recognition \cite{speech}. 
However, their inherent security issues have become increasingly prominent. One of the widely studied problems is the adversarial attack \cite{FGSM, zhou2024securely,zhou2023advclip,song2025segment}, where the adversary manipulates the model to output incorrect results by adding perturbations to the inputs. 
Previous works focus on adversarial examples (AEs) in the digital domain, which can be categorized into two approaches: perturbation-based methods \cite{pgd,zhou2024darksam}, and patch-based methods \cite{advpatch,copy-paste}. 
The former involve adding perturbations usually constrained by a specific norm, offering better stealthiness, while the latter involve applying elaborate patches to specific regions of the images, providing better attack performance but breaching the stealthiness.

\begin{figure}[t!]
   \setlength{\belowcaptionskip}{-1.5em}  
    \centering
    \includegraphics[width=0.48\textwidth]{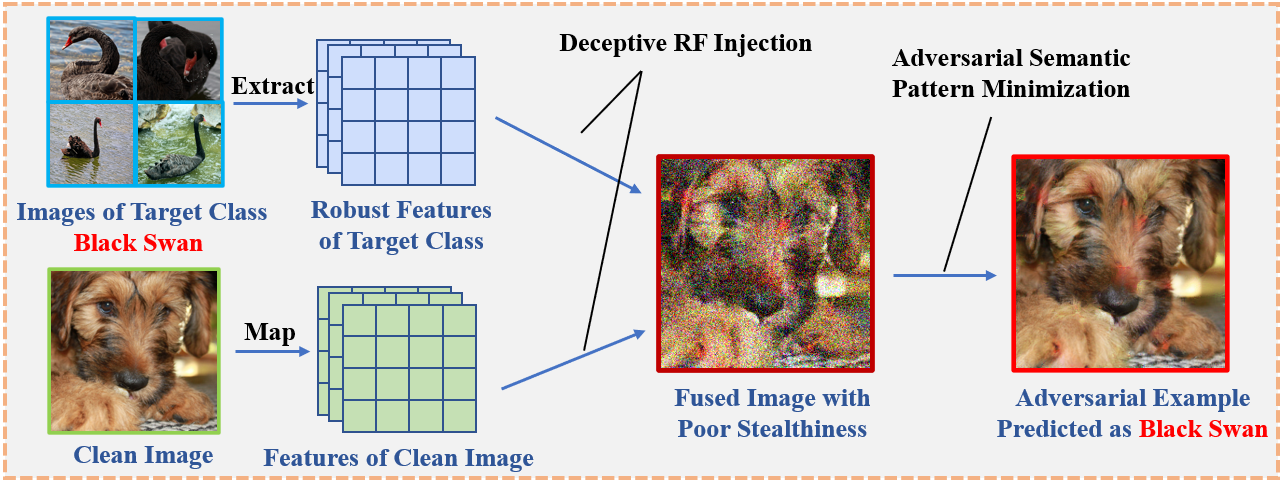}
    \caption{Our strategies Deceptive RF Injection and Adversarial Semantic Pattern Minimization.}
    \label{strategy}
\end{figure}

With the application of DNNs in the physical world, such as autonomous driving \cite{zhang2024detector} and facial recognition \cite{li2024transfer}, recent works pay more attention to AEs in real-world scenarios, known as physical-world adversarial examples (PAEs). 
Due to {environmental factors} in the physical world (\eg, distance, angles, and lighting), only a few perturbation-based methods \cite{luzhaojun,hijacking} exhibit effective attacks in real-world scenarios, but they still suffer from poor transferability and lack robustness to {changes of environmental factors}. Current works on PAEs mainly focus on patch-based methods \cite{RP2, tpa,DOE} and optical-based methods \cite{advlb,shadowattack, rfla} that leverage beams or shadows in the physical world. 
However, both of them have {inherent limitations} \cite{phy_survey}. Patch-based methods compromises the stealthiness to maintain attack performance, making PAEs more detectable. Optical-based methods are fragile to {the variation of environmental factors}, limiting its effectiveness to specific scenarios. In summary, existing PAEs face two challenges: the first is the unsatisfactory attack performance in real-world scenarios, \ie, poor transferability and robustness, while the second is difficult trade-off between attack effectiveness and stealthiness.


Due to the inherent limitations of patch-based and optical-based methods, it is difficult to fundamentally address these challenges by them, so we explore a perturbation-based method to overcome the challenges.
{Recent works  \cite{rf1,rf2} point out that existing perturbation-based methods concentrate merely on adversarially manipulating non-robust features (N-RFs), which are highly predictive (\ie, playing a critical role in the model's prediction) but sensitive to perturbations and vary across models. Thus, the adversarial N-RFs fail to neither influence the model's prediction when faced with changes of environmental factors in real-world scenarios nor perceived by other black-box models \cite{rf5}, which arises the first challenge.}
In contrast, there exists another type of predictive features known as robust features (RFs). They are strongly correlated with the image's semantics, robust to perturbations, and can be perceived by different models \cite{rf_nature,rf_nature2}. 
Therefore, we propose a novel strategy to overcome the first challenge, referred as {Deceptive RF Injection}, which involves covering RFs of other classes onto the predictive features in clean images.
{For the second challenge, we propose another strategy, Adversarial Semantic Pattern Minimization, which involves removing most perturbations and preserving only essential adversarial semantic patterns in AEs. Our strategies are illustrated in Fig. \ref{strategy}.}

Based on above two strategies, we propose Robust Feature Coverage Attack (RFCoA) to generate PAEs with excellent transferability, robustness and stealthiness, which consists of Robust Feature Disentanglement and Adversarial Feature Fusion. In Robust Feature Disentanglement, we design an optimization process to extract RFs of the target class. During Adversarial Feature Fusion, we adopt the attention mechanism to fuse the RFs with clean images and optimize the attention weights of RFs to accurately covering predictive features in clean images, thus achieving targeted adversarial attack. Besides, according to the second strategy, we combine the minimal cognitive pattern approach \cite{cognition} to eliminate unnecessary perturbations and extract adversarial semantic patterns from fusion results by optimizing a pattern mask.

We evaluate our method on ImageNet ILSVRC 2012 \cite{imagenet} in both digtal and physical scenarios.
The experimental results demonstrate that our method outperforms all existing state-of-the-art (SOTA) physical-world adversarial attacks in terms of transferability, robustness, and stealthiness. 
{Furthermore, we also demonstrate the effectiveness of our method on large vision-language models (LVLMs)}~\cite{zhang2024badrobot,wang2024trojanrobot}, such as MiniGPT-4 \cite{minigpt} and LLaVA \cite{llava}, which indicates its potential in more complex scenarios and tasks.

In conclusion, the key contributions of our work are outlined as follows:
\begin{itemize}
    \item[1)] {We provide a comprehensive review and summary of the challenges of existing PAEs and propose two novel strategies, Deceptive RF Injection and Adversarial Semantic Pattern Minimization, to fundamentally address the challenges.}

    \item[2)] Based on the proposed two strategies, we design a novel physical-world adversarial attack method RFCoA with high transferability, robustness and stealthiness, which consists of Robust Feature Disentanglement and Adversarial Feature Fusion.

    \item[3)] Extensive experiments demonstrate the superiority of our method compared with existing SOTA methods. Additionally, we also demonstrate the effectiveness of our method on LVLMs, indicating its potential for applying to more complex scenarios.

\end{itemize}

\begin{table}[t]
\centering
\setlength{\belowcaptionskip}{-2em}
\setlength{\tabcolsep}{2pt}
\small
\resizebox{0.48\textwidth}{!} {
\begin{tabular}{c|ccccc}
\toprule[1.8pt]
\cellcolor[rgb]{.95,.95,.95} \textbf{Method} &
\cellcolor[rgb]{.95,.95,.95} \makecell{\textbf{Type}} &
\cellcolor[rgb]{.95,.95,.95} \makecell{\textbf{Knowledge}} &
\cellcolor[rgb]{.95,.95,.95} \makecell{\textbf{Transferability}} &
\cellcolor[rgb]{.95,.95,.95} \makecell{\textbf{Robustness}} &
\cellcolor[rgb]{.95,.95,.95} \makecell{\textbf{Stealthiness}} \\
\midrule[1.8pt]
\makecell{AdvPatch \cite{advpatch}} & \textbf{Patch} & \textbf{White-box} & \Circle & \LEFTcircle & \Circle \\
\makecell{TPA \cite{tpa}} & \textbf{Patch} & \textbf{Black-box} &  \LEFTcircle &  \CIRCLE & \Circle \\
\makecell{AdvLB \cite{advlb}} & \textbf{Optical} & \textbf{Black-box} & \Circle & \Circle & \CIRCLE \\
\makecell{TnTAttack \cite{tnt}} & \textbf{Patch} &\textbf{White-box} & \Circle & \CIRCLE & \Circle \\
\makecell{C/P Attack \cite{copy-paste}} & \textbf{Patch} & \textbf{Black-box}  & \LEFTcircle & \CIRCLE & \Circle \\
ShadowAttack~\cite{shadowattack} & \textbf{Optical} & \textbf{Black-box} & \LEFTcircle &  \Circle & \CIRCLE \\
RFLA \cite{rfla} & \textbf{Optical} & \textbf{Black-box} &  \LEFTcircle &  \Circle & \CIRCLE \\
CleanSheet~\cite{hijacking} & \makecell{ \textbf{Perturbation}} & \textbf{Black-box} & \LEFTcircle &  \LEFTcircle & \CIRCLE \\

\midrule[1.3pt]
\cellcolor[HTML]{FFF7F0} \textbf{RFCoA (Ours)} &\cellcolor[HTML]{FFF7F0} \makecell{\textbf{Perturbation}}  &\cellcolor[HTML]{FFF7F0} \textbf{Black-box} & \cellcolor[HTML]{FFF7F0} \CIRCLE &\cellcolor[HTML]{FFF7F0} \CIRCLE & \cellcolor[HTML]{FFF7F0} \CIRCLE \\
\bottomrule[1.8pt]
\end{tabular}
}
\caption{Comparison among existing representative works on PAE under image classification task and our method. {``\CIRCLE"} indicates that the method performs well in the aspect and {``\LEFTcircle"} indicates while the method shows some improvement in this aspect, it remains mediocre overall.}

\label{Tab:Comparison}
\end{table}

\section{Related Work}
\subsection{Adversarial Example in the Digital Domain}
Adversarial examples (AEs) \cite{FGSM, zhou2023downstream} are created by introducing imperceptible adversarial perturbations into images, leading to incorrect outputs of the model during the inference stage.
Existing works on AEs in digital domain mainly focus on perturbation-based methods, such as FGSM \cite{FGSM}, PGD \cite{pgd}, and C\&W \cite{CW}. These methods effectively attack white-box models but face challenges in transferring to black-box models. Furthermore, the introduced adversarial perturbations are fragile and tend to lose their effectiveness in real-world scenarios \cite{phy_survey,zhou2025numbod}. Other patch-based methods, like AdvPatch \cite{advpatch} and DPatch \cite{dpatch}, exhibit better robustness and have potential for application in the physical world, but the adversarial patches are too conspicuous, making them easy-to-detect.

\subsection{Adversarial Example in the Physical World}
More recently, an increasing number of works focus on deploying AEs in the physical world.
Due to the fragility of perturbation-based methods to environmental factors, only a few works (\eg, HA\&NTA\cite{luzhaojun} and CleanSheet \cite{hijacking}) are effective in real-world scenarios, but their transferability and robustness remain poor.
Some works (\eg, RP2 \cite{RP2}, Copy/Paste Attack \cite{copy-paste}, T-Sea \cite{t-sea}, DOE \cite{DOE}, and etc.) involve carefully designing patches or camouflages and applying adaptive transformations, like Expectation Over Transformation (EOT) \cite{EOT}, to enhance the robustness to physical-world perturbations. Despite achieving relatively good effectiveness on white-box models, they fail to remain satisfactory performance on black-box models. Moreover, they sacrifice the stealthiness of PAEs, making them more detectable, which is the inherent shortcoming of this kind of methods.
Others like ShadowAttack \cite{shadowattack}, AdvLB \cite{advlb}, and RFLA \cite{rfla} leverage optical perturbations like beam and shadows. While they exhibit excellent stealthiness, their inherent deficiency is the limited robustness, making them highly sensitive to the environments.

In summary, as shown in Tab. \ref{Tab:Comparison}, existing works on PAEs face two challenges: unsatisfactory attack performance in real-world scenarios (\ie, poor transferability and robustness), and difficulty in balancing attack effectiveness with stealthiness. Considering the inherent deficiencies of patch-based and optical-based methods, we design a perturbation-based method to overcome these challenges in this paper.

\begin{figure}[t!]
   \setlength{\belowcaptionskip}{-1.5em}  
    \centering
    \includegraphics[width=0.48\textwidth]{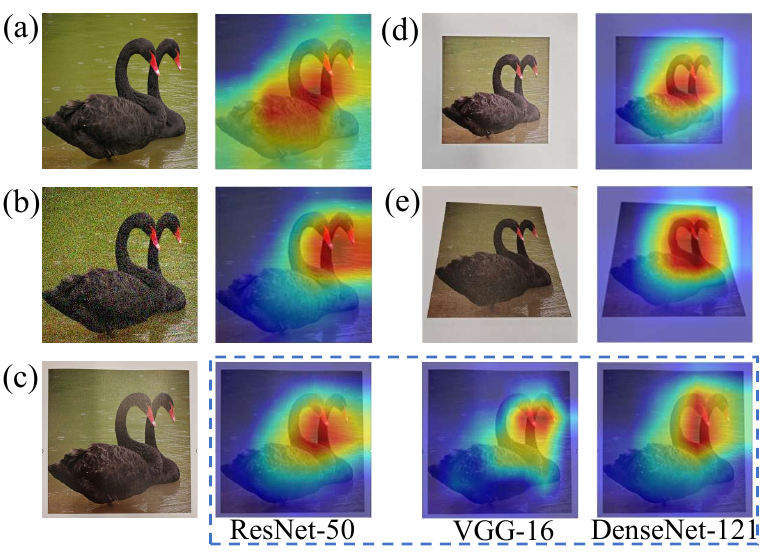}
    \caption{Attention maps calculated by Grad-CAM. (a) is the clean image in the digital world, (b) is the image added random noise in the digital world, and (c) to (e) are sampled in the physical world with various distances and angles. Notably, except for the special annotations, all models used to compute the attention maps are ResNet-50.}
    \label{rf_phy}
\end{figure}

\section{Methodology}

\begin{figure*}[ht!]
   \setlength{\belowcaptionskip}{-1em}  
    \centering
    \includegraphics[width=0.95\textwidth]{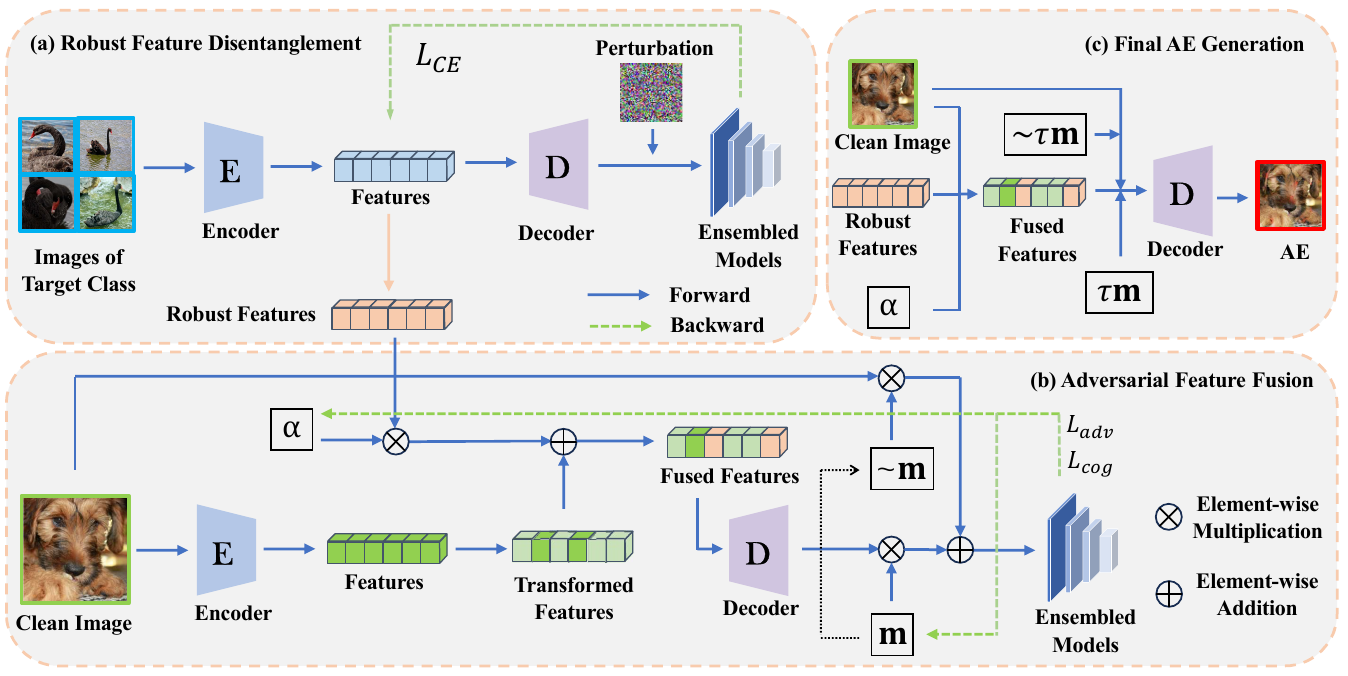}
    \caption{The overview of our method. (a) and (b) are the two modules of our method. After optimizing the $\alpha$ and $\mathbf{m}$ in (b), we calculate the final PAE by them through (c).}
    \label{overview}
\end{figure*}

\subsection{Problem Definition}
Notably, the attack we consider is in the black-box scenario, {where the adversary can only access the dataset information, the output of the victim model $f$ and cannot obtain the model's parameters or intermediate results. However, the adversary can employ several surrogate models with different structures from $f$ trained on the dataset.} This threat model is consistent with many existing works on PAEs \cite{DOE,t-sea}.

Considering that the perturbations from the physical world reduce the model accuracy, we choose to launch the targeted attack, which are more challenging than the untargeted attack. Given a classifier $f$, a clean image $x$ and its label $y$, and a target class $t$, The adversary's goal is to create an adversarial example $x'$ that satisfies Eq. (\ref{e1}):
\begin{equation}
\begin{aligned}
    & min \quad \|x' - x\|  \quad \\
    & {s.t.} \quad f(x') = t
\end{aligned}
\label{e1}
\end{equation}

\subsection{Intuition}
Recent works \cite{rf1,wangunlearnable} point out that AEs generated by existing perturbation-based methods merely focus on adversarially manipulating non-robust features (N-RFs) that are highly predictive but brittle to perturbations and vary across different models, to cause the model's incorrect outputs. 
Due to environmental perturbations in real-world scenarios, the adversarial effectiveness of these N-RFs is significantly degraded, failing to influence the model's prediction. Furthermore, adversarial N-RFs are also difficult for other black-box models to perceive, let alone manipulate their outputs. Therefore, existing perturbation-based methods on PAEs exhibit deficiencies in both robustness and transferability. 

\textbf{(i) How to fundamentally improve the transferability and robustness of AEs?}  We need to leverage another type of predictive features that are more robust to perturbations and be perceived by multiple models, to launch attack. Fortunately, robust features (RFs) happen to satisfy both of these requirements perfectly \cite{rf_nature,rf_nature2}. Although these properties have been demonstrated in the digital domain, further exploration in the physical world is still necessary.
Here, we sample images in both digital and physical worlds, input them into models, and utilize Grad-CAM \cite{gradcam} to visualize the attention maps, as shown in Fig. \ref{rf_phy}. Comparing (a) and (b), in the image perturbed by noise, the model's attention shift from the feather texture of the black swan to the neck, head, and beak, indicating that RFs are distributed in these regions. According to (c) to (e), the model's attention maps largely overlap with those in (b), demonstrating RFs remain highly predictive and robust to changes of environmental conditions in the physical world. Additionally, in Fig. \ref{rf_phy} (c), different models exhibit highly similar attention patterns for the physical-world sample, all concentrating on RFs. The above observation indicates the properties of RFs also valid in the physical world. Thus, we can utilize them to fundamentally improve the transferability and robustness of AEs.

\textbf{(ii) How to achieve better trade-off between the attack performance and stealthiness?} Due to the perceptibility of RFs to humans, directly integrating them into clean images would compromise the stealthiness of AEs. Therefore, we dynamically adjust the weights of RFs during the fusion process and eliminate unnecessary perturbations to improve the stealthiness by minimizing adversarial patterns.



\subsection{Overview of RFCoA}
The overview of our method, RFCoA is shown in Fig. \ref{overview}. It consists of two stages: Robust Feature Disentanglement and Adversarial Feature Fusion. In the first stage, we extract the RFs of the target class. In the second stage, we cover these RFs onto the clean image's predictive features. During this process, we optimize the weights $\alpha$ of the RFs in the fusion and the pattern mask $\mathbf{m}$ to minimize the visual difference between the AE and the clean image while maintaining attack performance. Finally, we execute the fusion process with the optimized $\alpha$ and $\mathbf{m}$ to generate the final AE, incorporating transparency $\tau$ to further improve stealthiness.

\subsection{Robust Feature Disentanglement}
Formally, the definition of robust features is as Eq. (\ref{rf}) :
\begin{equation}
    \mathbb{E}_{(x,y) \in \mathcal{D}} \left[ \inf_{\delta \in \Delta(x)} f \left( y \cdot E(x + \delta) \right) \right] \geq \gamma
\label{rf}
\end{equation}

where $x$ and $y$ are images and labels in the dataset $\mathcal{D}$, $\delta$ is the perturbation constrained in $\Delta(x)$, $E$ is the robust feature extractor, $f (\cdot)$ is a function used to evaluate the correlation between features and labels, and $\gamma$ is a predefined threshold. 

The above definition implies that even affected by noise, the RFs in images remain highly predictive, allowing the model to rely on them for accurate predictions. In light of this, we design an optimization-based method to extract RFs. First, we employ the encoder of a pre-trained autoencoder to map images of the target class into the feature space to initialize the optimization target.
{Then, we reconstruct these features into images by the corresponding decoder, add noise, and input them into the model. After calculating the cross-entropy loss with the label, we iteratively optimize these features.}
Moreover, to enhance the universality of the extracted features, enabling them perceivable by models with different architecture, we ensemble several models and calculate the average loss.

Specifically, we formally express the optimization process for Robust Feature Disentanglement as Eq. (\ref{e3}):

\begin{equation}
    \begin{aligned}
f_{t} = \operatorname*{arg\,min}_{f} \quad &  \frac{1}{N} \sum_{i=1}^{N} L_{CE}(\mathcal{M}_{i}( \mathbf{D}(f)  + \delta ), y_{t}) \\
\text{s.t.} \quad & f_{0} = \mathbf{E}(x), \quad \|\delta \|_{\infty} \leq \epsilon 
\end{aligned}
\label{e3}
\end{equation}
where $\mathbf{E}$ and $\mathbf{D}$ are the encoder and decoder of the pre-trained autoencoder, $x$ and $y_{t}$ are images and label in the target class, $L_{CE}$ represents the cross-entropy loss, $N$ is the number of ensembled models, and $\mathcal{M}_{i}$ is the $i$-th model. Notably, $f_{0}$ is the initial value of the optimization target $f$, and the infinity norm of the perturbation $\delta$ is constrained by $\epsilon$.

\subsection{Adversarial Feature Fusion}
After extracting  RFs of the target class, we then fuse them into clean images. 
In this stage, we ensures the attack performance in two aspects: weakening the predictive features in clean images, and overlaying RFs of the target class into original images. Due to the uneven distribution of predictive features in clean images, the weights of features at different positions should vary during fusion. Therefore, we adopt the attention mechanism \cite{attention}. Initially, for clean images, we compute the spatial attention map in the feature space, which can reflect the distribution of predictive features. To simplify and clarify our method, we draw inspiration from the Grad-CAM, which employs the magnitude of gradients to assess the importance of features for prediction~\cite{zhou2024darksam}. Specifically, the process to obtain spatial attention maps can be expressed as Eq. (\ref{e4}) :

\begin{equation}
    \mathcal{S} = F(|\nabla_{f_{c}} \frac{1}{N} \sum_{i=1}^{N}  L(\mathcal{M}_{i}(\mathbf{D}(f_{c}), y) )|)
\label{e4}
\end{equation}
where $f_{c}$ is the representation of the clean image $x_{c}$ in the feature space, \ie, $\mathbf{E}(x_{c})$, $F$ is the sigmoid function that maps the absolute values of the gradients to range [0,1], and $\mathcal{S}$ is the calculated spatial attention map with the same shape as $f_{c}$. 
Given that positions with higher values in the attention map indicate concentrated predictive features, we transform the clean image's features as described in Eq. (\ref{e5}), thereby weakening the influence of predictive features of $x_{c}$ in the subsequent fusion process.

\begin{equation}
    f'_{c} = (1 - \mathcal{S}) \circ f_{c}
    \label{e5}
\end{equation}

{To accurately cover RFs of the target class onto predictive features of clean images distributed at different positions, we also employ the attention mechanism for the RFs in the fusion process, as shown in Eq. (\ref{e6}).} $x'$ is the fused image and $\alpha$ is the attention weights of RFs. Then we optimize $\alpha$ by minimizing the adversarial loss defined in Eq. (\ref{e7}).

\begin{equation}
    x' = \mathbf{D}( \alpha \circ f_{t} + f'_{c}) 
    \label{e6}
\end{equation}

\vspace{-1em}
\begin{equation}
    L_{adv} = \frac{1}{N} \sum_{i=1}^{N} (w_{1} L(\mathbf{M}_{i} (x'), y_{t}) - w_{2} L(\mathbf{M}_{i}(x'), y_{c}))
    \label{e7}
\end{equation}
where $w_{1}$ and $w_{2}$ are pre-set weight parameters.
The first term aims to make the fused image exhibit the semantics of $f_{t}$ as much as possible, while the second term further weakening the influence of $f_{c}$.

However, the above fusion process does not consider the trade-off between the attack performance and stealthiness of the generated AEs. Here, we employ the minimal cognitive pattern \cite{cognition} to remove unnecessary corruption and preserve only essential adversarial patterns. The fusion process should be rewritten as:
\begin{equation}
    x' = \mathbf{m} \circ \mathbf{D}( \alpha \circ f_{t} + f'_{c}) + ( 1- \mathbf{m}) \circ x_{c} 
    \label{e8}
\end{equation}
Notably, $\mathbf{m}$ is the pattern mask with the same shape as $x_{c}$ and needs to be optimized along with $\alpha$ during the fusion process. Additionally, we introduce the cognitive loss $L_{cog}$ to constrain the corruption to the clean image.
\begin{equation}
    L_{cog} = w_{3} \| \mathbf{m} \|_{1}  + w_{4} TV(\textbf{m}) - w_{5} SSIM(x',x)
    \label{e9}
\end{equation}
where $\| \cdot \|_{1}$ is the $l_{1}$ norm, $TV(\cdot)$ represents the total variation loss, $SSIM(\cdot)$ is the Structural Similarity Index Measure (SSIM) to measure the similarity between $x'$ and $x_{c}$, and $w_{3}$ to $w_{5}$ are pre-set weights. The first item aims to eliminate unnecessary perturbations in the fusion results by minimizing the norm of pattern mask, while the latter two terms ensure the generated AEs have a smoother visual appearance and a higher similarity to the original image.

In conclusion, the whole optimization during the fusion can be expressed as Eq. (\ref{e10}) :
\begin{equation}
    \operatorname*{arg\,min}_{\alpha, \mathbf{m}} \quad L_{adv} + L_{cog}
    \label{e10}
\end{equation}
Furthermore, to facilitate direct control over the visual appearance of the AEs, we introduce a transparency parameter, $\tau$, which serves as a weight factor on $\mathbf{m}$ when calculating the final result by Eq. (\ref{e8}).


\section{Experiments}
In this section, we evaluate our method in both digital and physical worlds and compare it with existing SOTA works on PAEs. In addition, we also discuss the attack performance on LVLMs and defenses strategies. The ablation studies are provided in the supplementary.

\begin{table*}[t]
  \centering
  \setlength{\tabcolsep}{8pt}
   \setlength{\belowcaptionskip}{-1em}  
\begin{tabular}{ccccccccccccc}
    \toprule[1.5pt]
    \multirow{2}{*}{\diagbox{Attacks}{Models}} & \multicolumn{2}{c}{\textbf{RN-50*}} & \multicolumn{2}{c}{RN-101} & \multicolumn{2}{c}{WRN-50} & \multicolumn{2}{c}{WRN-101} & \multicolumn{2}{c}{\textbf{VGG-16*}} & \multicolumn{2}{c}{VGG-19} \\
          & Dig. & Phy. & Dig. & Phy. & Dig. & Phy. & Dig. & Phy. & Dig. & Phy. & Dig. & Phy. \\
    \midrule
    Clean Acc.    & 0.76  & 0.60  & 0.77  & 0.62  & 0.79  & 0.64  & 0.79  & 0.68  & 0.72  & 0.56  & 0.72  & 0.61 \\
    C/P-A & 0.88  & 0.37 & 0.12  & 0.07 & 0.27  & 0.20 & 0.14  & 0.08 & 0.78  & 0.49 & 0.62  & 0.37 \\
    TPA   & 0.88  & 0.41 & 0.10  & 0.05 & 0.12  & 0.11 & 0.07  & 0.05 & 0.95  & 0.55 & 0.65  & 0.39 \\
    RFLA  & 0.87  & 0.44 & 0.25  & 0.15 & 0.29  & 0.21 & 0.26  & 0.18 & 0.90  & 0.48 & 0.43  & 0.27 \\
    CS & 0.84  & 0.46 & 0.51  & 0.22 & 0.33  & 0.19 & 0.18  & 0.11 & \textbf{0.99} & 0.51 & 0.89  & 0.35 \\
    \rowcolor[rgb]{.95,.95,.95}
    Ours  & \textbf{1.00} & \textbf{0.87} & \textbf{0.59} & \textbf{0.38} & \textbf{0.65} & \textbf{0.60} & \textbf{0.43} & \textbf{0.39} & \textbf{0.99} & \textbf{0.92} & \textbf{0.96} & \textbf{0.67} \\
    \midrule[1.5pt]
    \multirow{2}{*}{\diagbox{Attacks}{Models}} & \multicolumn{2}{c}{\textbf{DN-121*}} & \multicolumn{2}{c}{DN-169} & \multicolumn{2}{c}{DN-201} & \multicolumn{2}{c}{ShuffleNet-v2} & \multicolumn{2}{c}{ViT-b32} & \multicolumn{2}{c}{GoogleNet} \\
          & Dig. & Phy. & Dig. & Phy. & Dig. & Phy. & Dig. & Phy. & Dig. & Phy. & Dig. & Phy. \\
    \midrule
    Clean Acc.    & 0.74  & 0.58  & 0.76  & 0.63  & 0.77  & 0.65  & 0.76  & 0.66  & 0.76  & 0.63  & 0.70  & 0.59 \\
   C/P-A & 0.69  & 0.32 & 0.17  & 0.13 & 0.20  & 0.14 & 0.17  & 0.12 & 0.04  & 0.02 & 0.16  & 0.10 \\
    TPA   & 0.93  & 0.40 & 0.15  & 0.10 & 0.06  & 0.04 & 0.07  & 0.06 & 0.08  & 0.05 & 0.03  & 0.03 \\
    RFLA  & 0.83  & 0.43 & 0.39  & 0.27 & 0.23  & 0.16 & 0.14  & 0.07 & 0.06  & 0.03 & 0.17  & 0.12 \\
    CS & 0.98  & 0.42 & 0.71  & 0.36 & 0.68  & 0.31 & 0.17  & 0.11 & 0.04  & 0.02 & 0.11  & 0.08 \\
     \rowcolor[rgb]{.95,.95,.95}
    Ours  & \textbf{0.99} & \textbf{0.71} & \textbf{0.74} & \textbf{0.69} & \textbf{0.79} & \textbf{0.59} & \textbf{0.40} & \textbf{0.28} & \textbf{0.18} & \textbf{0.14} & \textbf{0.30} & \textbf{0.23} \\
    \bottomrule[1.5pt]
  \end{tabular}%
   \caption{Quantitative results of attacks across various models in both digital and physical worlds. {``Clean Acc."} values denote the clean accuracy, while the other rows report tASR values. Notably, the models marked with {``*"} denote our surrogate models that are treated as white-box models, and the others are black-box models.}
  \label{digital}
\end{table*}

\begin{figure*}[ht]
   \setlength{\belowcaptionskip}{-1em}  
    \centering
    \includegraphics[scale=0.78]{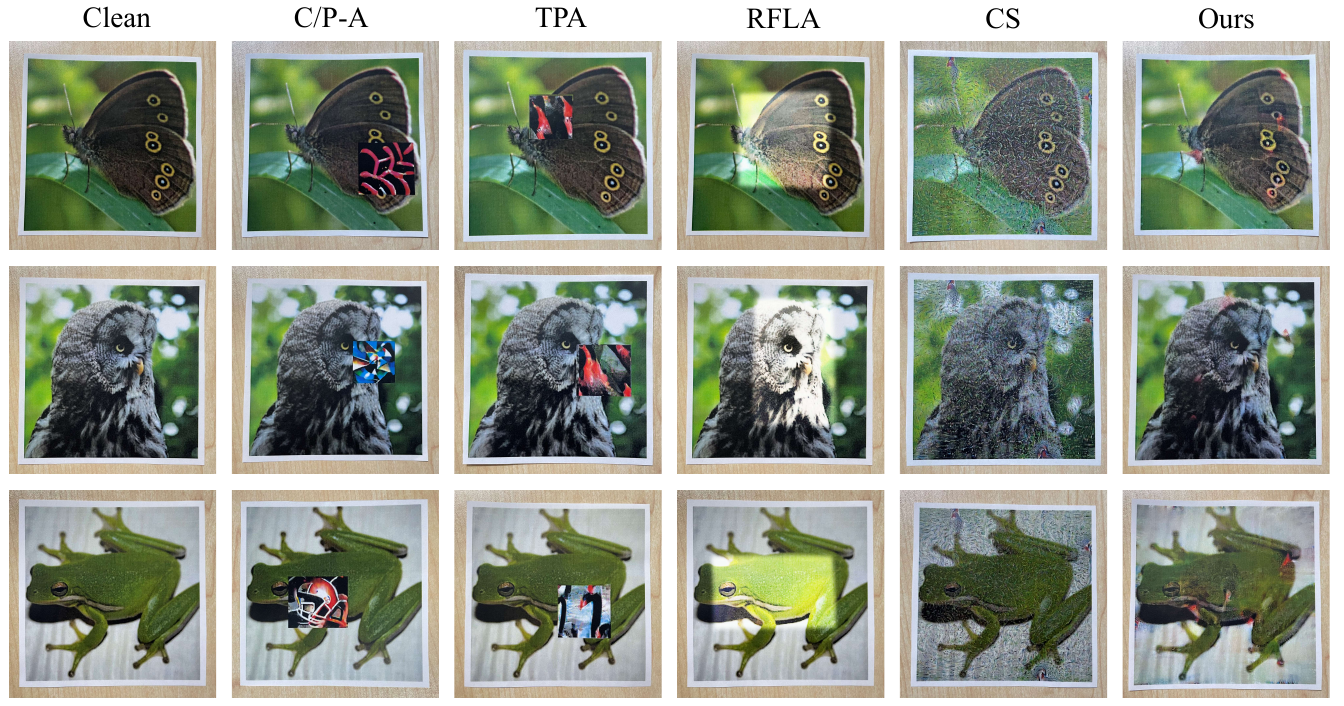}
    \caption{Visualization results of PAEs in the physical world.}
    \label{phy_vis}
\end{figure*}

\subsection{Setup}
\textbf{Dataset and classifiers.} We conduct experiments on ImageNet ILSVRC 2012 \cite{imagenet}. In preprocessing, we resize and crop the images to 224x224. To comprehensively evaluate transferability, we select 12 commonly used models in image classification, including the ResNet (RN) series \cite{resnet}, Wide ResNet (WRN) series \cite{wrn}, VGG series \cite{vgg}, DenseNet (DN) series \cite{densenet}, GoogleNet \cite{googlenet}, ShuffleNet \cite{shufflenet}, and Vision Transformer (ViT) \cite{attention}. For each attack, we use three surrogate models: ResNet-50, VGG-16, and DenseNet-121, treating the others as black-box models.

\bigskip

\noindent\textbf{Attack settings.} We compare our method with existing SOTA physical-world adversarial attacks in image classification, including TPA \cite{tpa}, Copy/Paste Attack (C/P-A) \cite{copy-paste}, RFLA \cite{rfla} and CleanSheet (CS) \cite{hijacking}. We utilize the official open-source code and select the settings claimed to achieve the best performance for evaluation. Details of our method's settings can be found in the supplementary.



\bigskip

\noindent\textbf{Metrics.}
We adopt four metrics for evaluation: Clean Accuracy, Target Attack Success Rate (tASR)~\cite{zhangdenial}, Structural Similarity Index Measure (SSIM) \cite{ssim}, and Learned Perceptual Image Patch Similarity (LPIPS) \cite{lpips}. Clean Accuracy measures the model's accuracy on clean inputs, while tASR represents the rate at which adversarial examples are misclassified as the target class. SSIM and LPIPS evaluate the stealthiness of AEs, with a smaller LPIPS value and a larger SSIM value suggesting better stealthiness.



\subsection{Attack Performance}
\noindent\textbf{Sampling.} In the digital world, we randomly select 1000 images from ImageNet and set the attack target for each image to a randomly chosen class other than the original label. 
In the physical world, we randomly select 100 generated AEs, print them on 10cm x 10cm white paper, and photograph each with an iPhone 14 from a 10cm distance. The images are then resized to 224x224 for model inputs.The tASR results across different models are reported in Tab. \ref{digital}.

\bigskip


As shown in Tab. \ref{digital}, these attacks exhibit high tASR values on white-box models but vary significantly in black-box scenarios. Patch-based methods C/P Attack and TPA show poor transferability, with low tASR on most black-box models. RFLA performs relatively well on WRN series models but has limited transferability to other black-box models. CleanSheet shows improved transferability in the digital world, but suffers significant reduction on tASR values in physical-world scenarios. In contrast, our method achieves the highest tASR across all black-box models, including those with distinct architectures such as ViT-b32 and GoogleNet, demonstrating superior effectiveness and transferability in both digital and physical worlds.

\subsection{Robustness and Stealthiness}
\noindent\textbf{Evaluation of robustness.}
Here, we consider two common perturbation factors in the physical world: the distance and the angle during sampling. To this end, we set three sampling distances of 10cm, 15cm, and 20cm, and three angles of 15\degree, 30\degree, and 45\degree. The results are presented in Tab. \ref{stealthiness}.

\begin{table}[t!]
  \centering
   \setlength{\belowcaptionskip}{-1em}  
  \resizebox{0.5\textwidth}{!}{%
    \begin{tabular}{ccccccc}
    \toprule[1.5pt]
    \multirow{2}{*}{\diagbox{Attacks}{Distances}} & \multicolumn{2}{c}{10cm} & \multicolumn{2}{c}{15cm} & \multicolumn{2}{c}{20cm} \\
          & W-b   & B-b   & W-b   & B-b   & W-b   & B-b \\
    \midrule
    C/P-A & 0.41  & 0.14  & 0.36  & 0.13  & 0.30  & 0.09  \\
    TPA   & 0.45  & 0.10  & 0.41  & 0.09  & 0.36  & 0.07  \\
    RFLA  & 0.45  & 0.16  & 0.38  & 0.12  & 0.34  & 0.10  \\
    CS & 0.46  & 0.19  & 0.43  & 0.16  & 0.37  & 0.12  \\
    \rowcolor[rgb]{.95,.95,.95}
    Ours  & \textbf{0.83} & \textbf{0.44} & \textbf{0.78} & \textbf{0.41} & \textbf{0.74} & \textbf{0.38} \\
    \midrule[1.5pt]
    \multirow{2}{*}{\diagbox{Attacks}{Angles}} & \multicolumn{2}{c}{15\degree} & \multicolumn{2}{c}{30\degree} & \multicolumn{2}{c}{45\degree} \\
          & W-b   & B-b   & W-b   & B-b   & W-b   & B-b \\
    \midrule
    C/P-A & 0.41  & 0.13  & 0.35  & 0.11  & 0.30  & 0.08  \\
    TPA   & 0.41  & 0.08  & 0.35  & 0.06  & 0.32  & 0.05  \\
    RFLA  & 0.41  & 0.13  & 0.27  & 0.06  & 0.19  & 0.03  \\
    CS & 0.42  & 0.18  & 0.36  & 0.13  & 0.31  & 0.09  \\
    \rowcolor[rgb]{.95,.95,.95}
    Ours  & \textbf{0.79} & \textbf{0.39} & \textbf{0.73} & \textbf{0.35} & \textbf{0.56} & \textbf{0.22} \\
    \bottomrule[1.5pt]
    \end{tabular}%
  }
   \caption{Comparative results of average tASR under various distance and angles. {``W-b"} denotes white-box models, while {``B-b"} represents black-box models.}
  \label{sample}
\end{table}

\begin{table}[t!]
  \centering
   \setlength{\belowcaptionskip}{-2em}  
   \resizebox{0.4\textwidth}{!}{
  \begin{tabular}{cccccc}
    \toprule[1.5pt]
    Methods & C/P-A & TPA   & RFLA  & CS & Ours \\
    \midrule
    SSIM ($\uparrow$)  & 0.56  & 0.68  & 0.84  & 0.77  & \textbf{0.89} \\
    LPIPS ($\downarrow$) & 0.59  & 0.36  & 0.20  & 0.25   & \textbf{0.14} \\
    \bottomrule[1.5pt]
  \end{tabular}%
  }
   \caption{Comparative results of average SSIM and LPIPS values across different methods.}
  \label{stealthiness}
\end{table}


From Tab. \ref{sample}, RFLA and CleanSheet are sensitive to variations in sampling distance and angles, showing significant degradation in tASR and indicating limited robustness in the physical world. In contrast, the patch-based attacks, C/P Attack and TPA, exhibit better robustness, with minimal changes in tASR. Our method demonstrates the highest robustness, maintaining a tASR of 0.22 on black-box models even under the challenging 45\degree sampling condition, significantly outperforming the other methods.

\bigskip

\noindent\textbf{Evaluation of stealthiness.} We select the AEs in Section 4.2 and compute their SSIM and LPIPS values relative to clean images in the digital domain to evaluate their stealthiness. The experimental results are presented in Tab. \ref{stealthiness} and some physical-wrold visualization results are shown in Fig. \ref{phy_vis}.

For C/P Attack and TPA, due to the use of patches that significantly differ from the original images, AEs are easily perceived by humans, with SSIM values below 0.7. In contrast, RFLA, CleanSheet, and our method introduce mild perturbations, avoiding visually disruptive areas and thus achieving better stealthiness. Notably, our method also outperforms RFLA and CleanSheet in numerical results, demonstrating its superior stealthiness.

\subsection{Attack against LVLMs}
To explore the potential of our method on complex tasks and models, we apply AEs to LVLM-based Visual Question Answering (VQA) and image description tasks.

\textbf{Settings.} We employ 100 AEs generated by our method to launch targeted attack against MiniGPT-4 \cite{minigpt} and LLaVA \cite{llava} and record the average tASR values. Notably, the attack is considered successful if the LVLMs' response includes the target class or its synonyms.
Besides, we also enhance the prompts by appending the target class name or related terms at the end. The results are shown in Tab. \ref{vlm}. More visualization results and test screenshots are provided in the supplementary.

\begin{table}[t!]
  \centering
  \setlength{\belowcaptionskip}{-1.5em}
   \resizebox{0.5\textwidth}{!}{%
    \begin{tabular}{ccccc}
    \toprule[1.5pt]
   \multirow{2}{*}{\diagbox{Prompts}{Models}} & \multicolumn{2}{c}{MiniGPT-4} & \multicolumn{2}{c}{LLaVA} \\
          & VQA   & Description & VQA   & Description \\
    \midrule
    Normal & 0.36    & 0.44    & 0.32    & 0.41 \\
    Enhanced & 0.57    & 0.65    & 0.50    & 0.57 \\
    \bottomrule[1.5pt]
    \end{tabular}%
    }
    \caption{Quantitative results of our method on LVLMs.}
  \label{vlm}%
\end{table}%


As shown in Tab. \ref{vlm}, our method outperforms in image description tasks compared to VQA. This is likely because, in image description, the model's attention is spread across the entire image, increasing the chances of recognizing adversarially injected RFs. Additionally, incorporating the target class name or related terms in textual prompts significantly boosts tASR. We hypothesize that these enhanced prompts guide the LVLM to focus more on the RFs of the target class, resulting in outputs more closely aligned with it. In conclusion, these results demonstrate the potential of our attack method to transfer to LVLM and multimodal tasks.

\begin{figure}[t]   \centering 
\setlength{\abovecaptionskip}{-0.2em}
\setlength{\belowcaptionskip}{-2em}
\subfigure[PGD-AT]{\includegraphics[width=0.49\linewidth]{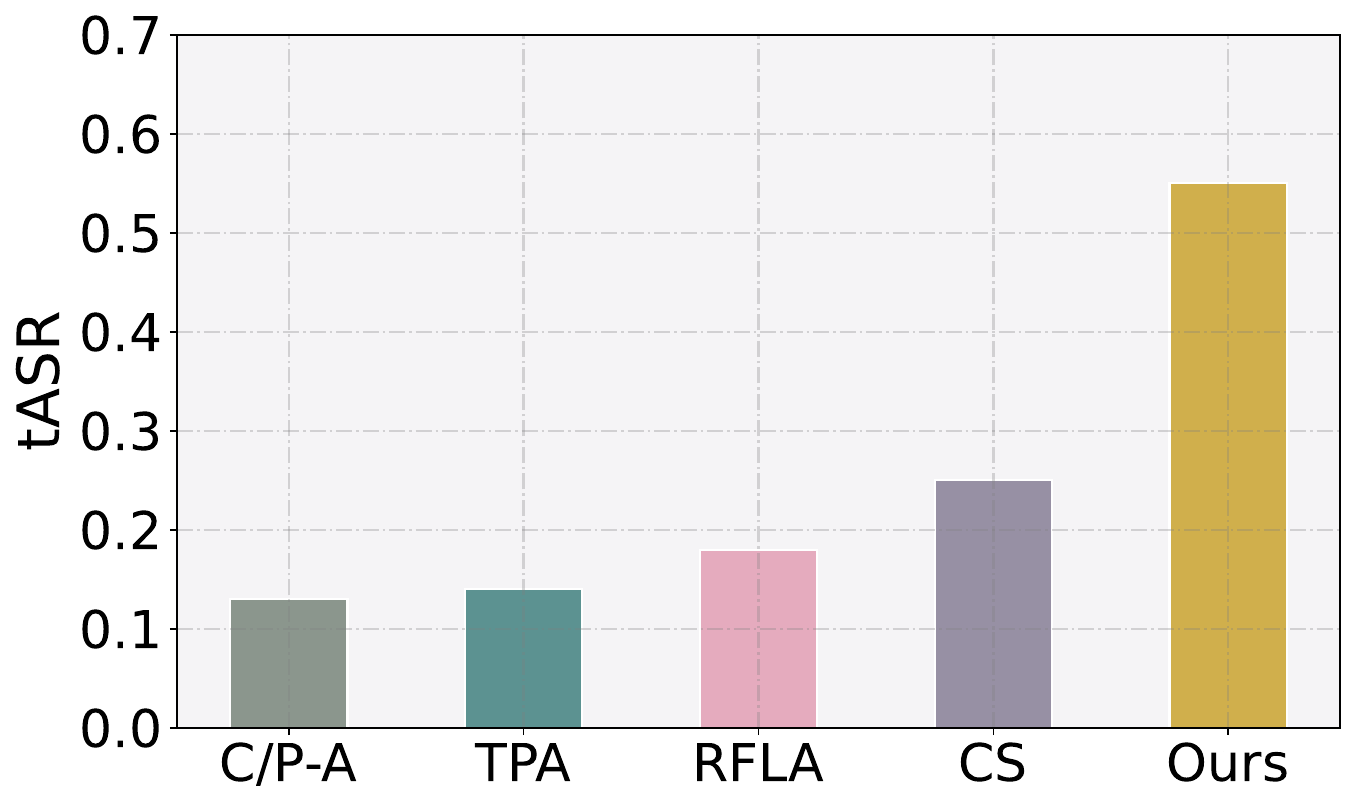}}       
\subfigure[DiffPure]{\includegraphics[width=0.49\linewidth]{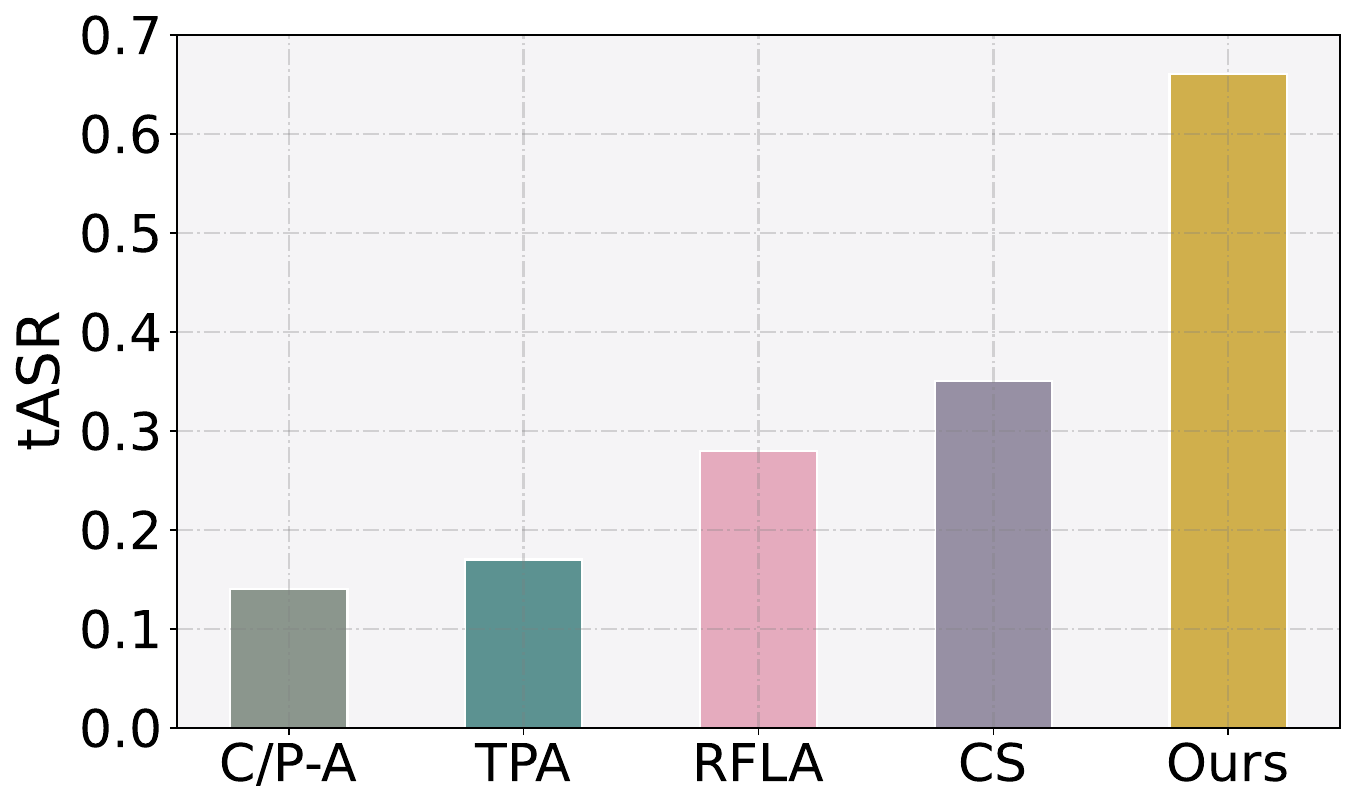}}            \caption{The average tASR of attacks on white-box models under defenses.}    \label{defense}   
\end{figure}

\subsection{Defense}
Here, we discuss attacks under defense measures. We consider two of the most common defense methods: adversarial training (AT) and purification, and evaluate in the digital world on white-box models. For the former, we employ PGD-AT \cite{pgd-at} for testing, with training epochs of 5 and perturbation budget of 16/255. For the latter, we choose Diffusion Purifier (DiffPure) \cite{diffpure}, keeping the parameter settings consistent with the original work.
The results are present in Fig. \ref{defense}.

According to the results, our method, compared to the others, maintains strong attack performance under both defenses, with an average tASR exceeding 50\%.  Specifically, AT reduces the model's sensitivity to non-robust features but does not affect the model's ability to perceive robust features \cite{rf1}. As a result, the models after AT can still recognize the robust features we inject. Moreover, due to the semantic nature of robust features, purification techniques are also unable to entirely eliminate them.

\section{Conclusion}
In this work, we propose two novel strageties Deceptive RF Injection and Adversarial Semantic Pattern Minimization to fundamentally overcome the challenges of existing PAEs. Based on the strategies, we design a perturbation-based attack RFCoA to craft PAEs with excellent transferability, robustness, and stealthiness. Experimental results demonstrate the superiority of our method compared with existing SOTA works in both digital and physical scenarios. Moreover, our method shows effectiveness on LVLMs, which indicates the potential of our attack to transfer to more complex tasks. 

\section*{Acknowledgments}
Minghui's work is supported by the National Natural Science Foundation of China (Grant No. 62202186). 
Shengshan's work is supported by the National Natural Science Foundation of China (Grant No. 62372196).
Wei Wan is the corresponding author.

\bibliography{main}

\end{document}